\pdfoutput=1

\documentclass[11pt]{article}

\usepackage[]{emnlp2021}

\usepackage{times}
\usepackage{latexsym}

\usepackage[T1]{fontenc}

\usepackage[utf8]{inputenc}

\usepackage{microtype}

\newcommand{\system}{{\tt CAT}\xspace}
\newcommand{\mixup}{{\tt CMIX}\xspace}

\usepackage{color}

\usepackage{mathtools}
\usepackage{mwe}
\usepackage{multirow}
\usepackage{amsfonts,amssymb, amsbsy}
\usepackage{subfig}
\usepackage{graphicx}
\usepackage{color}
\usepackage{booktabs}
\usepackage{amsmath}

\DeclareMathOperator*{\argmax}{arg\,max}
\newcommand{\norm}[1]{\left\lVert#1\right\rVert}
\usepackage[ruled,vlined]{algorithm2e}
\usepackage{mathrsfs}
\usepackage{tabularx}
\usepackage{graphicx}
\usepackage[normalem]{ulem}
\graphicspath{ {./images/} }
\usepackage[]{graphicx}
\usepackage{caption}
\usepackage{tikz}

\title{Counterfactual Adversarial Learning with Representation Interpolation}

\author{
Wei Wang$^1$, Boxin Wang$^2$, Ning Shi$^1$ \\
{\bf Jinfeng Li$^1$, Bingyu Zhu$^1$, Xiangyu Liu$^1$, Rong Zhang$^1$ } \\
$^1$Alibaba Group \\
$^2$University of Illinois at Urbana-Champaign \\
\textit{\{luyang.ww, shining.shi, jinfengli.ljf,} \\
\textit{zhubingyu.zby, eason.lxy, stone.zhangr\}@alibaba-inc.com} \\
\textit{boxinw2@illinois.edu} \\}

\begin{document}
\maketitle
\begin{abstract}
Deep learning models exhibit a preference for statistical fitting over logical reasoning. Spurious correlations might be memorized when there exists statistical bias in training data, which severely limits the model performance especially in small data scenarios. In this work, we introduce \textbf{C}ounterfactual \textbf{A}dversarial \textbf{T}raining framework (\system) to tackle the problem from a causality perspective. Particularly, for a specific sample, \system first generates a counterfactual representation through latent space interpolation in an adversarial manner, and then performs Counterfactual Risk Minimization (CRM) on each original-counterfactual pair to adjust sample-wise loss weight dynamically, which encourages the model to explore the true causal effect. Extensive experiments demonstrate that \system achieves substantial performance improvement over SOTA across different downstream tasks, including sentence classification, natural language inference and question answering. \footnote{Code is available at \url{https://github.com/ShiningLab/CAT.git}}
\end{abstract}

\section{Introduction}

Large-scale pre-trained language models such as BERT \citep{devlin2019bert}, as one of the recent breakthroughs, have revolutionized the model development paradigm in natural language processing (NLP) and improved traditional task-specific models by a large margin.
Although the pre-training and fine-tuning framework has been shown to be effective in transferring the pre-learned knowledge to downstream tasks and boosting the model performance, it could be a double-edged sword if there exists statistical bias in the training dataset, especially in small data scenarios \citep{yue2020interventional}. Taking sentiment analysis as an example, the downstream classifier can easily mistake a certain person name for a sentimental word if it is imbalanced distributed in positive and negative samples. \citet{yue2020interventional} further show that when the capacity of the model to certain semantics is strong, it will in turn strengthen this bias, in which case the large pre-trained model becomes an amplifier for spurious features. 

Generally, such problem can be better solved from the perspective of causality \citep{zhang2020causal}.
In the context of causal inference, causation is not correlation but something more essential with the mechanism of data generation \citep{pearl2009causal}. Statistical bias, also named spurious correlations, is a result of \textit{confounder}, a variable that can influence dependent variables and independent variables simultaneously \citep{pearl2009causality}. 
The study of causation aims to find the true causal effects between variables that help to unveil the true casual effect behind observation data and realize more robust inference. 

Specifically, counterfactuals are one feasible way to discover the causation. Counterfactual examples are defined as the ones that minimally-different from original ones but lead to different labels \citep{teney2020learning}. Recent work shows that counterfactual samples can significantly improve model generalization and boost model
performance. 
\citet{kaushik2019learning} and \citet{teney2020learning} use additional human-labeled or augmented counterfactual examples to mitigate spurious correlations.
\citet{zeng2020counterfactual} proposes a two-stage training approach for named entity recognition tasks for better generalization.

Although these methods gain significant improvements in model performance, they are limited in practice since they are either designed for specific tasks or requiring human-labeled samples. 
Additionally, no optimization 
is conducted on those 
counterfactual examples which could further improve the performance shown in our work.

\vspace{2pt} \noindent 
\textbf{Our work.} 
We revisit the above problem in NLP domain from a causality perspective. Following the definition of counterfactual examples, our motivation is to explore the following question: 
\textit{What would be the minimal intervention that alters the model output?} As a result, we propose \system, an end-to-end and task-agnostic \textbf{C}ounterfactual \textbf{A}dversarial \textbf{T}raining framework during fine-tuning to introduce counterfactual representations in training stage through latent space interpolation.

Concretely, to define our problem, we first introduce Structural Causal Model (SCM) to view our problem through a casual graph that depicts the data generation mechanism. 
To cut off the confounder, we conduct \textit{do-calculus} \citep{pearl1995causal} 
for causal effect adjustments. 
Specifically, we propose a counterfactual representation interpolation technique called \mixup which is a variant of \textit{Mixup} \citep{zhang2018mixup} to generate counterfactuals and approximately realize \textit{do-calculus} in deep learning framework. 
For each example $x$ in training set, \mixup samples a counterpart $x'$  and generates a counterfactual representation by interpolating the representation of $x$ and $x'$, which is adaptively optimized by a novel Counterfactual Adversarial Loss (CAL)  to minimize the differences from original ones but lead to drastic label change by definition. 
Finally, to connect each original-counterfactual pair, besides the traditional Empirical Risk Minimization (ERM) \citep{vapnik1998statistical}, We extend it to a new counterpart, i.e., Counterfactual Risk Minimization (CRM), to allow the model to adjust sample-wise loss weight dynamically so as to explore the causal effect behind data rather than simple correlations memorization. 

We also extend \system to other complicated tasks besides simple classification, which are rarely studied by other Mixup-based methods.
Through extensive experiments on text classification, natural language inference and question answering on two SOTA baselines, BERT and RoBERTa \citep{liu2019roberta}, we observe consistent improvement for \system in promoting the testing accuracy especially for small data in an extra-data-free manner.

Our contributions are summarized as follows:
\begin{itemize}
  \item We investigate the problem of spurious correlations from a causality perspective which has not been widely studied in conventional statistical learning.
  \item 
  We propose \mixup for counterfactual representation interpolation to approximate \textit{do-calculus} realization in deep learning framework, which is adaptively optimized by a novel Counterfactual Adversarial Loss. Moreover, we extend the traditional ERM to a novel Counterfactual Risk Minimization as a new learning principle connecting original data representations and counterfactual ones, which enables \system to explore causal effects and debias the spurious correlation.
    \item We propose \system as a general framework for various types of NLU tasks, including sentence classification, natural language inference and question answering tasks. We show that \system outperforms SOTA by a large margin across different tasks particularly when data is limited. 
\end{itemize}
    
\section{Related Work}
\vspace{2pt} \noindent 

\vspace{2pt} \noindent
\textbf{Large-scale Pre-trained Language Model.} The most widely-used solution for alleviating spurious correlations is using large-scale dataset. \citet{yang2019xlnet} and \citet{liu2019roberta} proposed to build the pre-trained language models utilizing even larger corpus to reduce the bias. 

\vspace{2pt} \noindent 
\textbf{Data Augmentation.}
Data augmentation is another solution and has become a de facto technique used in state-of-the-art machine learning models. 
\citet{zhang2015character} performed text augmentation by replacing words or phrases with their synonyms, and recently \citet{wei2019eda} proposed more operations.
Using word embedding, \cite{wang2015s}  tried to find a similar word for replacement.
In addition, back translations \cite{sennrich2016improving} and contextual augmentation \cite{fadaee2017data,kobayashi2018contextual} techniques have been proposed to replace target words. 

Through shrinking the weight of the training data relative to  L2 regularization, mixup \cite{zhang2018mixup} trained a neural network on convex combinations of pairs of examples and their labels to generate new samples $(\tilde{x}, \tilde{y})$ from $(x_i, y_i)$ and $(x_j, y_j)$, formally as 
\begin{equation}
\begin{aligned}
\tilde{x} & = \lambda x_i + (1-\lambda)x_j, \\
\tilde{y} & = \lambda y_i + (1-\lambda)y_j.
\end{aligned}
\end{equation}
Followed by mixup, more works \cite{verma2019manifold,verma2019interpolation,verma2019graphmix,berthelot2019mixmatch,yun2019cutmix} are proposed, mainly focusing on image-format data. Recently, such regularization techniques were brought into NLP tasks \citet{chen2020mixtext}. 

\vspace{2pt} \noindent 
\textbf{Adversarial Training.} 
Adversarial training has been proven to be an effective approach for improving the robustness of neural network models ~\cite{DBLP:conf/iclr/MiyatoDG17,madry2018towards,tramer2017ensemble,shrivastava2017learning}.
Specifically, \citet{DBLP:conf/iclr/MiyatoDG17} applied adversarial and virtual adversarial training to text domain by applying perturbations to the word embeddings, which achieved state-of-the-art results on multiple semi-supervised and purely supervised tasks. 
By minimizing the resultant adversarial risk inside different regions around input samples, \citet{zhu2019freelb} proposed a novel adversarial training algorithm, FreeLB, that is capable of promoting higher invariance in the embedding space. 

\vspace{2pt} \noindent 
\textbf{Causal Inference.}
\citet{yue2020interventional} explored estimating true causal effects on few-shot images classification, which is an intervention-based approach. Similarly, \cite{tang2020long} focuses on long-tail data set. 
In addition, using counterfacutals is another way to discover the causation in training data. For example, \citet{kaushik2019learning} and \citet{teney2020learning} leveraged additional human-labeled counterfactual examples, while \citet{zeng2020counterfactual} proposed a two-stage training approach for named entity recognition task. These studies have been proved that counterfactual samples are more valuable in boosting model performance compared to normal samples.


\begin{figure*}[tb]
 \centering
 \setlength{\abovecaptionskip}{3pt}
 \includegraphics[scale=0.7]{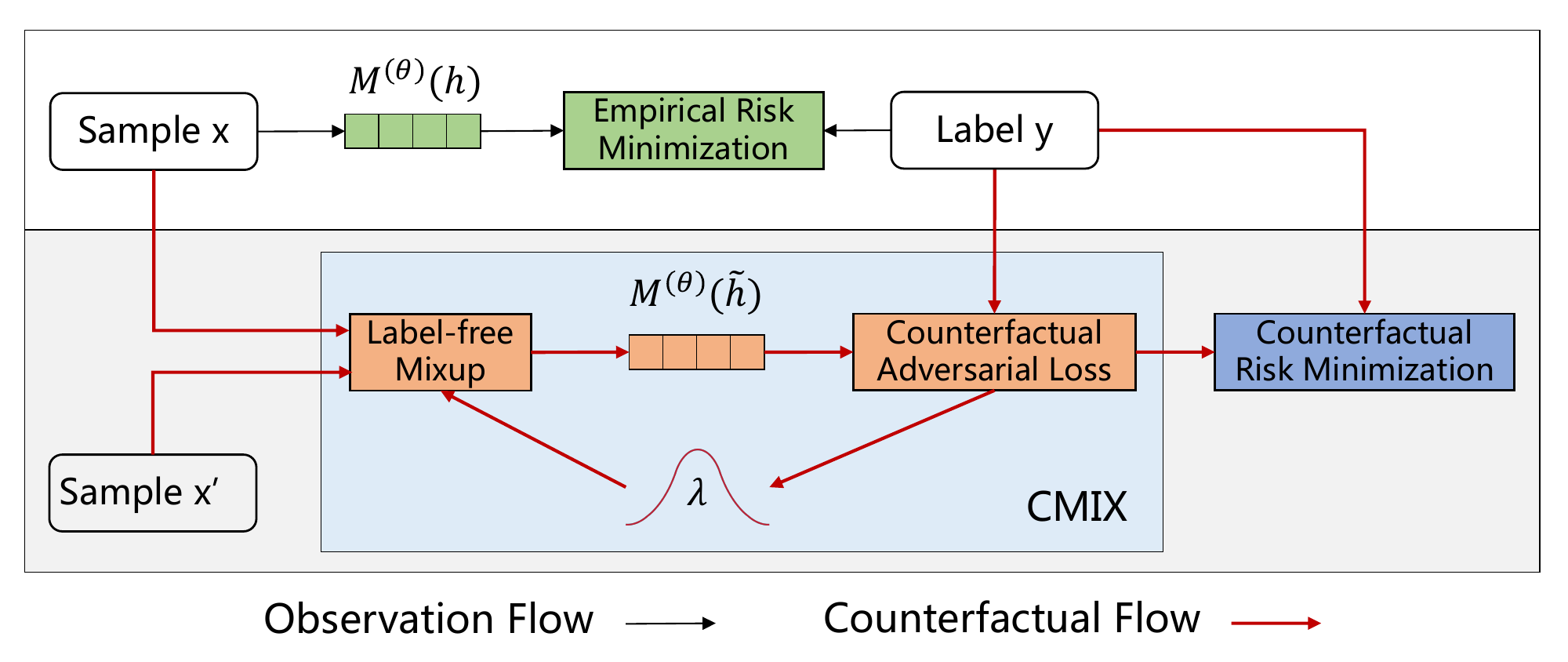}
 \caption{The framework of \system. Besides the normal supervised ERM (Observation) flow on the top, for a certain observation $x$, \system will randomly sample another $x'$ from training data. Then a counterfactual representation $\tilde{h}$ is generated and optimized by \mixup.  Finally, CRM is applied on final model output $M^{(\theta)}(\tilde{h})$.}
 \label{fig:framework}
 \vspace{-0.35cm}
\end{figure*}

\section{CAT}
\label{sec:length}
In this section, we introduce how \system solves the problem of spurious correlations from a causal perspective (see Figure \ref{fig:framework}). We first explicate our problem by introducing Structural Causal Model (SCM). Then we illustrate our approach by answering the following questions: (1) How can we cut off the confounder for spurious bias elimination by counterfactual representations? (2) How can we optimize our counterfactual representations? (3) How can we learn from both original and counterfactual representations to debias the spurious correlations?

\subsection{Problem Definition} \label{SCM}


To explicate our problem, we first introduce SCM to depict the data generation mechanism. Every SCM can be represented as a directed acyclic graph (DAG) which can be written as:
\begin{equation}
\begin{aligned}
\small \mathbf{X}_i  \coloneqq f_i(\mathbf{X}_{pa(i)}) + \mathbf{U}_i, i =1, 2,..., d,
\end{aligned}
\end{equation}
where each $\mathbf{X}_i$ is an endogenous variable in the graph, $\mathbf{X}_{pa(i)}$ denotes the set of parent variables of $\mathbf{X}_i$, $\mathbf{U}_i$ denotes independent and identically distributed random noise,  $d$ is the number of variables, and $f_i(\mathbf{X}_{pa(i)})$ represents the direct causation from $\mathbf{X}_{pa(i)}$ to $\mathbf{X}_i$. Each $\mathbf{U}_i$ is called an exogenous variable because it is determined outside the graph.

\begin{figure}[htb]
 \centering
 \setlength{\abovecaptionskip}{3pt}
 \includegraphics[scale=0.35]{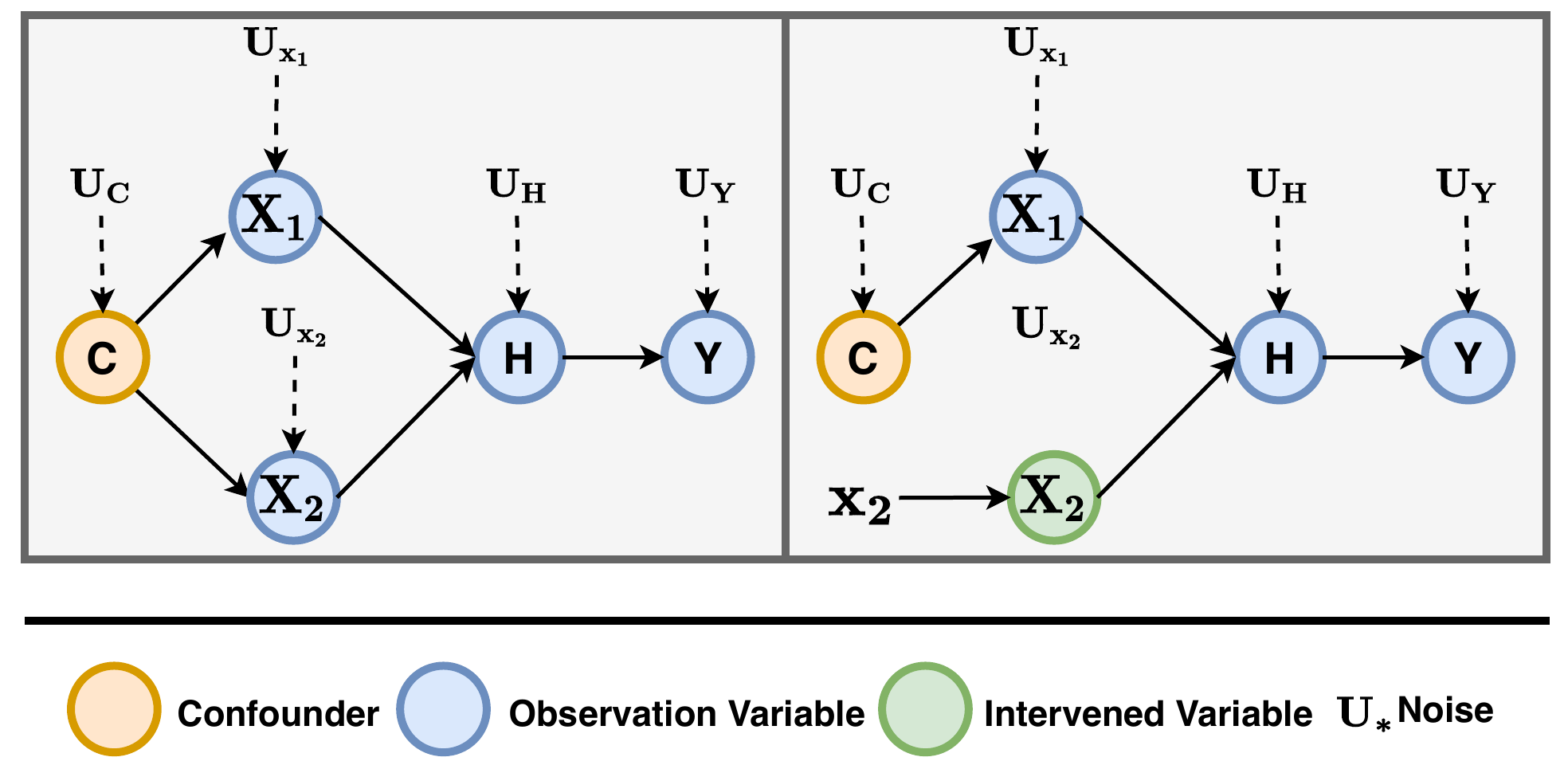}
  \caption{SCM of data generation mechanism. Left: Spurious correlations exist between $\textbf{X}_1$ and $\textbf{X}_2$ in observation data caused by confounder $\textbf{C}$. Right: Confounder is eliminated by \textit{do-calculus}.}
  \label{fig:SCM}
  \vspace{-0.3cm}
\end{figure}

We can use SCM to describe how our training data is generated and where spurious correlations are derived from fine-tuning process.
As shown in the left part of Figure \ref{fig:SCM}, $\mathbf{C}$ is the confounding variable which is the common cause of samples $\mathbf{X}_1$ and $\mathbf{X}_2$ and leads to spurious correlations between them. 
During fine-tuning, model $\mathbf{H}$ can easily take such training-data-specific spurious correlations as features to predict $\mathbf{Y}$, which could severely undermine model performance when inferring on test set where such correlations do not hold.
In practice, $\mathbf{C}$ can be subjective bias of human annotator, the domain of data, the region where data is collected, etc.

To eliminate such spurious correlations, we conduct \textit{do-calculus} on $\mathbf{X}_2$ (denoted as $do(\mathbf{X}_2)$), which is shown in the right side of Figure \ref{fig:SCM}. This operation is realized by cutting off all edges directing to $\mathbf{X}_2$ and setting $\mathbf{X}_2$ to a certain constant $\mathbf{x}_2$ (green node in Figure \ref{fig:SCM}). \textit{Do-calculus} blocks the causal effect from $\mathbf{C}$ to $\mathbf{X}_2$ so that $\mathbf{C}$ is no longer a confounder. In general, 
the operation allows us to estimate the true causal effect $P(\mathbf{Y}|do(\mathbf{X}))$ instead of the correlation $P(\mathbf{Y}|\mathbf{X})$ which is the conventional machine learning objective. 


\subsection{CMIX}
\textit{Do-calculus} is a  statistical tool derived from SCM for causal effect adjustments.
Recent study \citep{zeng2020counterfactual} proposed to use \textit{Counterfactuals} as an approximate realization of \textit{do-calculus} for neural networks. 
In this paper, instead of generating counterfactual examples that may require expensive human-annotation \citep{teney2020learning}, we propose \mixup as a variant of mixup \citep{zhang2018mixup} to generate counterfactual representations. 

\subsubsection{Label-free Mixup}
Due to the discrete nature of textual data, it is not favored to do interpolation directly on words as mixup does on images. Given that, \mixup conducts \textit{do-calculus} and generates counterfactual representations by interpolating the hidden states. We aim to set a new value for $\mathbf{X}$ in SCM independent of $\textbf{C}$ so that $\mathbf{X}$ is not affected by the \textit{confounder}. Unlike mixup-based methods \citep{chen2020mixtext} that require interpolation on labels, \mixup interpolates sample representations without label information, allowing \mixup to utilize unlabeled data.

Specifically, interpolation occurs on the hidden representations from the multi-layer transformers \citep{vaswani2017attention}. Formally, we denote the multi-layer transformer model by $M^{(\theta)}$ that takes the latent representation $h$ of $x$ as input and output the confidence prediction $y=M^{(\theta)}(h)$. Specifically, we denote the model $M^{(\theta)}$ at the $l$-th layer $M^{(\theta)}_l(\cdot), l\in\{0,1,...L\}$, then for two samples $x^{(i)}$ and $x^{(j)}$, their corresponding latent representations $h^{(i)}, h^{(j)}$ at the $l$-th layer are 
\begin{equation}
\label{eqn:repre}
\begin{aligned}
    h^{(i)}_0 &= W x^{(i)},  \quad h^{(j)}_0 = W x^{(j)}, \\
    h^{(i)}_l &= M^{(\theta)}_l(h^{(i)}_{l-1}), \quad l \in \{1,...,m\},\\
    h^{(j)}_l &= M^{(\theta)}_l(h^{(j)}_{l-1}), \quad l \in \{1,...,m\};
\end{aligned}
\end{equation}
where $W$ is the embedding look-up matrix to map the discrete sentence $x^{(i)}$ to the latent embedding $h_0$ as the $0$-th layer representation.
 Then we generate the counterfactual representation $\tilde{h}^{(i)}$ by  interpolation in $m$-th layer:
\begin{equation}
\label{eqn:Tmix}
\begin{aligned}
    &\tilde{h}^{(i)}_m = \lambda^{(i)} h_m^{(j)} + (1-\lambda^{(i)}) h_m^{(i)} \\
    &\tilde{h}^{(i)}_l = M^{(\theta)}_l(\tilde{h}^{(i)}_{l-1}), \quad l \in \{m+1,..., L\},
\end{aligned}
\end{equation}
where  $\lambda^{(i)}$ samples from beta distribution  $Beta(\alpha, \beta)$. We discuss the choice of beta distribution in Appendices \ref{appendix: beta_pdf}.

\vspace{2pt}
\noindent
\textbf{Disscussion About Attention Mask} \label{att_mask}: 
To address the attention mask of $\tilde{h}$ in BERT and its derivatives, which is not considered in \citep{chen2020mixtext},  we propose several possible ways to handle the attention mask of $\tilde{h}$, including using the mask of $h^{(i)}$, $h^{(j)}$ or using the last hidden layer as the interpolation layer to avoid the use of attention mask. After experiments, we find that using the mask of $h^{(i)}$ always achieves the best performance.

\subsubsection{Counterfactual Adversarial Loss}
We propose Counterfactual Adversarial Loss (CAL) to further optimize the counterfactual representations $\tilde{h}$ generated by \mixup. Following the definition of \citet{kaushik2019learning},  CAL optimizes counterfactual representations so that they are minimally-different from the original ones $x^{(i)}$ but lead to different labels $y^{(i)}$.
Specifically, we optimize the mixup parameter $\lambda^{(i)}$ by the following objective:
\[
\begin{minipage}{\displaywidth}\small
\noindent\begin{equation}
\begin{aligned}
\label{eqn:CFL_1}
	 \argmax_{\lambda^{(i)}} -\norm{\lambda^{(i)}}_p + \gamma L(M^{(\theta)}(\tilde{h}^{(i)}), y^{(i)}) 
	 \\ + \eta \Phi(M^{(\theta)}(\tilde{h}^{(i)})),
\end{aligned}
\end{equation}
\end{minipage}
\]
where $\norm{\cdot}_p$ is the $L^p$ norm, $L(\cdot,\cdot)$ is the loss function and  $\gamma$ and $\eta$ are the hyperparameters. $\Phi(\cdot)$ indicates extracting the maximum probability of a discrete distribution. 

CAL is a trade-off game for minimizing difference and maximizing label change. Maximizing $-\norm{\lambda^{(i)}}_p$ encourages smaller shifts of counterfactual representations, while maximizing loss function adversarially changes model prediction to anyone but not original label $y^{(i)}$. The last term is to make the model more confident about counterfactual representation predictions.

\begin{algorithm}[t]\small
    \caption{Counterfactual Adversarial Training Approach (\system)} \label{algos}
    \textbf{Input}: Dataset $D=\{(x^{(i)}, y^{(i)})\}_{i=1}^N$, model $M^{(\theta)}$, mixup layer candidate set $\mathcal{Q}$, Beta distribution parameters $\alpha$ and $\beta$, denote as $Beta(\alpha, \beta)$, couterfactual adversarial loss iteration step $\mathcal{L}$, warm up step $\mathcal{K}$, max step $\mathcal{T}$\\
    \For{step k $\in \{0,1,..., \mathcal{K}\}$}{
      Sample one batch $X^{(k)} \in D$. Denote corresponding representations as $h^{(k)}$;
      Do ERM on $M^{(\theta)}(h^{(k)})$\;
      }
    \For{step t $\in \{0,1,..., \mathcal{T\}}$}{
        Sample one batch $X^{(t)}$. Denote corresponding representations as $h^{(t)}$\;
        For each $x^{(i)}$ in $X^{(t)}$, random sample $q\in \mathcal{Q}$ and $\lambda^{(i)} \sim Beta(\alpha, \beta)$ and generate mixed representations in latent space using (Eq.\ref{eqn:Tmix}) \ to get one batch of counterfactual representations $\tilde{h}^{(t)}$\;
        \For{l $\in \{0,1,... \mathcal{L}\}$}{
            Optimize counterfactual representations using CAL (Eq.\ref{eqn:CFL_1})\;
        }
        Do CRM on $M^{(\theta)}(\tilde{h}^{(t)})$ and $M^{(\theta)}(h^{(t)})$\;
        Do ERM on $M^{(\theta)}(h^{(t)})$;
      }
\end{algorithm}

\subsection{Counterfactual Risk Minimization} \label{sec:CRM}
CRM is designed to enable the model to learn from both original representations and counterfactual ones. Recall that in supervised learning, given $(h,y)$ and their joint distribution $P$, the model $M^{\theta}: H \rightarrow Y$ is learnt by minimizing the average of loss function over data distribution $P$, also known as the \textit{expected risk}:
\begin{equation} \label{eqn:Erisk}
    R(M^{(\theta)}) = \int L(M^{(\theta)}(h), y)dP(h,y).
\end{equation}
Unfortunately, distribution $P$ in unknown in most practical situations so we approximate Eq.\ref{eqn:Erisk} by an empirical form:
\begin{equation} \label{eqn:emRisk}
     \hat{R}(M^{(\theta)}) = \frac{1}{n} \sum_{i=1}^n L(M^{(\theta)}(h^{(i)}), y^{(i)}).
\end{equation}
Minimizing Eq.(\ref{eqn:emRisk}) is known as the Empirical Risk Minimization (ERM) principle \citep{vapnik1998statistical}, which is widely adopted in most of machine learning models today.

Counterfactual Risk Minimization (CRM) is derived from ERM. Similar to \citet{swaminathan2015counterfactual}, \citet{charles2013counterfactual} and \citet{jung2020learning}, we rewrite Eq.(\ref{eqn:Erisk}): 
%
\[
\begin{minipage}{\displaywidth}\small
\noindent\begin{equation}\label{eqn:CRM}
\begin{aligned}
        R(M^{(\theta)}) &= \mathbb{E}_{h\sim P({H})}\mathbb{E}_{y\sim P({Y}|h)} L(M^{(\theta)}(h), y) \\
        &= \mathbb{E}_{h\sim P({H})}\mathbb{E}_{y\sim P({Y}|\tilde{h})} \left [ \frac{P(y|h)}{P(y|\tilde{h})}  L(M^{(\theta)}(h), y) \right ] \\ 
        &:= R_c(M^{(\theta)}),
    \end{aligned}
\end{equation}
\end{minipage}
\]
%
where the subscript $c$ denotes that $R_c(\cdot)$ is from the counterfactual distribution. Since we can estimate $P(y|h)$ and $P(y|\tilde{h})$ by the model $M^{(\theta)}$ , we can derive a tractable estimation for Eq.(\ref{eqn:CRM}) via Monte Carlo approximation:
\[
\begin{minipage}{\displaywidth}\small
\noindent\begin{equation}\label{eqn:CRM_1}
\begin{aligned}
    \hat{R}_c(M^{(\theta)}) &= \frac{1}{n} \sum_{i=1}^n \frac{\Phi (M^{(\theta)}(h^{(i)}))}{\Phi (M^{(\theta)}(\tilde{h}^{(i)}))} L(M^{(\theta)}(h^{(i)}), y^{(i)}) \\
    &= \frac{1}{n} \sum_{i=1}^n \hat{\omega}(h^{(i)}) L(M^{(\theta)}(h^{(i)}), y^{(i)}).
    \end{aligned}
\end{equation}
\end{minipage}
\]
We call Eq.(\ref{eqn:CRM_1}) \textit{counterfactual risk}, which can be also viewed as an importance sampling estimator that connects original-counterfactual distributions. Intuitively, CRM adjusts sample-wise loss weight dynamically according to $\hat{\omega}(h^{(i)})$.
The counterfactuals that have low confidence are more penalized and conversely the over-confident original data are discouraged. This
makes the decision boundary more discriminative and smooth. In practice, we bound $\hat{\omega}(h^{(i)})$ for numerical stability:
\[
\begin{minipage}{\displaywidth}\small
\noindent\
\begin{equation}\label{eqn:CRM_2}
    \hat{R}_c(M^{(\theta)}) = \frac{1}{n} \sum_{i=1}^n B(\hat{\omega}(h^{(i)})) L(M^{(\theta)}(h^{(i)}), y^{(i)}),
\end{equation}
\end{minipage}
\]
where
\begin{equation}
B(x)=\left\{
	\begin{aligned}
    & x \quad x \in [A_1, A_2]\\
	& A_1 \quad x < A_1\\
    & A_2 \quad x > A_2,\\
	\end{aligned}
	\right.
\end{equation}
for some $0<A_1<A_2$. 

During training, we first train a warm-up phase using only ERM like traditional fine-tuning. Then we start counterfactual adversarial training with ERM and CRM alternatively for succeeding steps. The whole framework is in algorithm \ref{algos}. One of the special cases are discussed in the \ref{appendix: same_label}. 
\vspace{2pt} \noindent 
\textbf{How CAL helps CRM}: (a) \underline{Stability:} In Eq.(\ref{eqn:CRM_1}), when the denominator $\Phi( M^{(\theta)}(\tilde{h}^{(i)}))$ gets close to zero, $\hat{R}_c(M^{(\theta)})$ can be arbitrarily far away from the true risk. The last term in CAL tend to maximize $M^{(\theta)}(\tilde{h}^{(i)})$ so that CRM is more stable. (b) \underline{Smaller variance:} The optimal choice is $P^{*}(y|\tilde{h}^{(i)}) = P(y|h^{(i)})  L(M^{(\theta)}(h)^{(i)}, y^{(i)})/\mu$ \citep{glasserman2004monte}, where $\mu$ is the true expected risk. In practice, although this is untraceable since $\mu$ is unknown, a $P(y|\tilde{h}^{(i)})$ approximately proportional to $P(y|h^{(i)})  L(M^{(\theta)}(h^{(i)}), y^{(i)})$ is preferred for variance reduction. 
Note that we can view traditional ERM as an importance sampling with $P(y|\tilde{h}^{(i)})=P(y|h^{(i)})$ and are estimated by $\Phi( M^{(\theta)}(h^{(i)}))$.
Here we consider two situations:

(1) When $L(M^{(\theta)}(h^{(i)}), y^{(i)})$ is large, then $\Phi(M^{(\theta)}(h^{(i)}))$ is likely small. This is an indication of the model sensitivity, which means the gradient regarding $ \gamma L(M^{(\theta)}(\tilde{h^{(i)}}), y^{(i)})$ in CAL is large to push the counterfactual in $\nabla \max ||\lambda^{(i)}||_p$ direction (towards $h^{(j)}$), hence to generate $\Phi(M^{(\theta)}(\tilde{h}^{(i)})) > \Phi(M^{(\theta)}(h^{(i)}))$ during optimization.

(2) When $L(M^{(\theta)}(h^{(i)}), y^{(i)})$ is small, which indicate the model is confident, gradient regarding $\lambda^{(i)}$ in $-\norm{\lambda^{(i)}}_p$ is more possible to dominate the total gradient of CAL and pull the counterfactuals close to $h^{(i)}$, then $\Phi(M^{(\theta)}(\tilde{h}^{(i)})) \approx \Phi( M^{(\theta)}(h^{(i)}))$. To conclude, in both situations,  $\Phi(M^{(\theta)}(\tilde{h}^{(i)}))$ is a better choice of $P(y|\tilde{h}^{(i)})$ (at least not worse) in the context of variance reduction.

\begin{table*}[t]
\centering
\resizebox{\textwidth}{!}{

\begin{tabular}{l}
    \hline
    \multirow{2}{*}{\textbf{Model}} \\
    \\
    \hline
    BERT\textsubscript{\begin{tiny}BASE\end{tiny}} \\
    TMix \\
    \system* \\
    \system \\
    \hline
    RoBERTa\textsubscript{\begin{tiny}BASE\end{tiny}} \\
    \system* \\
    \system \\
    \hline
    BERT\textsubscript{\begin{tiny}LARGE\end{tiny}} \\
    \system* \\
    \system \\
    \hline
    RoBERTa\textsubscript{\begin{tiny}LARGE\end{tiny}} \\
    \system* \\
    \system \\
    \hline
\end{tabular}

\begin{tabular}{cccc}
    \hline
    \multicolumn{4}{c}{\textbf{Yahoo! Answers}} \\
    \cline{1-4}
    \textbf{10} & \textbf{50} & \textbf{250} & \textbf{1000} \\
    \hline
    61.02 & 66.39 & 70.07 & 72.33 \\
    62.19 & 67.01 & 70.15 & 72.30 \\
    62.34 & 67.20 & 70.11 & 72.29 \\
    \textbf{63.53}& \textbf{68.11} &\textbf{71.40} &\textbf{72.52}   \\ 
    \hline
    61.95 & 66.96 & 69.61 & 71.21 \\
    63.09 & \textbf{67.84} & 70.08 & 71.95 \\
    \textbf{63.55} & 67.78 & \textbf{70.45} & \textbf{72.02} \\
    \hline
    63.54 & 67.96 & 70.75 & 72.93 \\
    64.33 & 68.07 & 70.72 & 72.95 \\
    \textbf{64.73} & \textbf{68.15} & \textbf{70.95} & \textbf{73.06} \\
    \hline
    64.38 & 67.80 & 70.60 & 72.28 \\
    66.20 & 68.92 & 71.10 & 72.90 \\
    \textbf{66.30} & \textbf{69.28} & \textbf{71.25} & \textbf{73.30} \\
    \hline
\end{tabular}

\begin{tabular}{cccc}
    \hline
    \multicolumn{4}{c}{\textbf{IMDB}} \\
    \cline{1-4}
    \textbf{10} & \textbf{50} & \textbf{250} & \textbf{1000} \\
    \hline
    73.28 & 78.03 & 82.38 & 85.88 \\
    74.32 & 78.64 & 82.58 & 85.90 \\
    73.77 & 78.98 & 82.45 & 85.96 \\
    \textbf{75.55} &\textbf{80.13} &\textbf{83.15} &\textbf{86.11} \\
    \hline
    81.57 & 84.30 & 87.00 & 88.36 \\
    82.80 & 85.11 & 87.40 & 88.45 \\
    \textbf{83.25} & \textbf{85.12} & \textbf{87.50} & \textbf{88.93} \\
    \hline
    76.51 & 81.22 & 85.42 & \textbf{87.32} \\
    \textbf{76.97} & 81.05 & 85.38 & 86.93 \\
    75.10 & \textbf{82.52} & \textbf{86.02} & 87.00 \\
    \hline
    81.50 & 87.63 & 89.03 & 90.06 \\
    79.95 & 87.55 & 89.48 & 90.10 \\
    \textbf{84.80} & \textbf{88.55} & \textbf{89.85} & \textbf{90.10} \\
    \hline
\end{tabular}

\begin{tabular}{cccc}
    \hline
    \multicolumn{4}{c}{\textbf{SNLI}} \\
    \cline{1-4}
    \textbf{10} & \textbf{50} & \textbf{250} & \textbf{1000} \\
    \hline
    42.68 & 57.62 & 70.17 & 77.16 \\
    43.90 & 58.55 & 70.57 & 77.40 \\
    44.37 & 59.42 & 71.23 & 77.89 \\
    \textbf{46.23} &\textbf{60.27} &\textbf{72.13} &\textbf{78.20}  \\
    \hline
    40.72 & 59.92 & 77.96 & 83.09 \\
    \textbf{41.95} & 63.33 & 79.15 & 83.25 \\
    41.30 & \textbf{64.47} & \textbf{79.69} & \textbf{83.75} \\
    \hline
    \textbf{44.33 }& 60.10 & 74.02 & 81.04 \\
    43.07 & 62.80 & 75.97 & 81.18 \\
    43.83 & \textbf{64.77} & \textbf{76.77} & \textbf{81.67} \\
    \hline
    38.22 & 62.73 & 82.27 & 85.99 \\
    39.15 & 61.85 & 82.90 & 85.63 \\
    \textbf{40.33} & \textbf{65.07} & \textbf{83.15} & \textbf{86.05} \\
    \hline
\end{tabular}

}
\caption{The average accuracy after multiple runs on Yahoo! Answers, IMDB and SNLI datasets. Bellowing the individual dataset is the number of training samples per class.}
\label{tab:main_results_tab1}
\end{table*}

\begin{table*}[t]
\centering
\resizebox{\textwidth}{!}{

\begin{tabular}{l}
    \hline
    \multirow{2}{*}{\textbf{Model}} \\
    \\
    \hline
    BERT\textsubscript{\begin{tiny}BASE\end{tiny}} \\
    \system* \\
    \system \\
    \hline
    BERT\textsubscript{\begin{tiny}LARGE\end{tiny}} \\
    \system* \\
    \system \\
    \hline
\end{tabular}

\begin{tabular}{ccc}
    \hline
    \multicolumn{3}{c}{\textbf{SQuAD 1.1}} \\
    \cline{1-3}
    \textbf{1/20} & \textbf{1/10} & \textbf{1/5} \\
    \hline
    51.83/62.50 & 66.06/76.56 & 72.25/81.75 \\
    \textbf{63.90/74.93} & 69.36/79.44 & 74.10/83.34 \\
    62.71/74.14 & \textbf{69.49/79.44} & \textbf{74.33/83.43} \\
    \hline
    70.66/81.29 & 75.85/85.16 & 79.14/87.24 \\
    72.18/82.15 & 75.69/84.83 & 79.06/87.08 \\
    \textbf{72.30/82.17} & \textbf{76.37/85.09} & \textbf{79.18/87.28} \\
    \hline
\end{tabular}

\begin{tabular}{ccc}
    \hline
    \multicolumn{3}{c}{\textbf{SQuAD 2.0}} \\
    \cline{1-3}
    \textbf{1/20} & \textbf{1/10} & \textbf{1/5} \\
    \hline
    51.10/54.12 & 55.60/58.84 & 61.84/65.42 \\
    55.44/57.55 & \textbf{59.84/62.44} & 61.77/64.97 \\
    \textbf{56.22/58.47} & 59.71/62.44 & \textbf{63.26/66.72} \\
    \hline
    59.41/63.03 & 66.28/70.30 & \textbf{71.30/74.88} \\
    61.84/65.27 & 66.55/70.08 & 69.40/72.87 \\
    \textbf{61.82/65.32} & \textbf{67.38/70.79} & 69.31/72.37 \\
    \hline
\end{tabular}

}
\caption{The model performance of EM/F1 on SQuAD 1.1 and SQuAD 2.0. Bellowing the individual dataset is the proportion of full training data used.}
\label{tab:qa_result_tab1}
\end{table*}

\begin{table}[t]
    \centering
    \resizebox{\linewidth}{!}{
    \begin{tabular}{l}
        \hline
        \multirow{2}{*}{\textbf{Model}} \\
        \\
        \hline
        BERT\textsubscript{\begin{tiny}BASE\end{tiny}} \\
        \system \\
        \hline
    \end{tabular}
    \begin{tabular}{cc}
        \hline
        \multicolumn{2}{c}{\textbf{SQuAD 1.1}} \\
        \cline{1-2}
        \textbf{EM} & \textbf{F1} \\
        \hline
        80.80 & 88.50 \\
        \textbf{81.77} & \textbf{88.98} \\
        \hline
    \end{tabular}
    \begin{tabular}{cc}
        \hline
        \multicolumn{2}{c}{\textbf{SQuAD 2.0}} \\
        \cline{1-2}
        \textbf{EM} & \textbf{F1} \\
        \hline
        72.57 & 75.99 \\
        \textbf{74.13} & \textbf{77.36} \\
        \hline
    \end{tabular}
    }
    \caption{The EM/F1 on full QA data.}
    \label{tab:qa_all_result}
\end{table}

\subsection{CAT in Question Answering} \label{adjust}
In addition to classification tasks that most mixup-based methods are examined on, we further extend \system to question answering where logic reasoning is preferred over correlation memorization. We propose four mixup strategies to handle question answering: i) directly mix, which is the same as in sentence classification task; ii) mix only on context; iii) mix only on queries; iv) mix only on non-answer contexts. Empirical results suggest that mixing only on non-answer contexts leads to the best and most consistent outcome. Regarding CAL and CRM, we sum up the start position loss and end position loss as the final loss.

\section{Experiments}
We evaluate \system on five widely used open-source benchmark datasets, including text classification, natural language inference and question answering.
\subsection{Dataset}

\vspace{2pt} \noindent 
\textbf{Yahoo! Answers} \citep{chang2008importance} consists of questions and their corresponding answers along with the categories that are assigned to questions. We carry out the same pre-processing as in \citep{chen2020mixtext}. 

\vspace{2pt} \noindent 
\textbf{IMDB} \citep{lin2011proceedings} is a typical dataset for binary sentiment analysis including 50k samples. 



\vspace{2pt} \noindent 
\textbf{SNLI} \cite{bowman-etal-2015-large} is a popular text entailment dataset that contains 570k human annotated sentence pairs.

\vspace{2pt} \noindent 
\textbf{SQuAD 1.1} \citep{rajpurkar-etal-2016-squad} consists of 100k question/answer pairs. Given a question
and a Wikipedia passage containing the answer, the task is to predict the answer span in the passage.

\vspace{2pt} \noindent 
\textbf{SQuAD 2.0} (\citep{rajpurkar-etal-2018-know}) combines the existing SQuAD 1.1 data with over 50k unanswerable questions written \textbf{adversarially} by crowdworkers.

For each dataset, experiments are conducted on multiple data sizes.  
Experiments are controlled in an incremental manner where we gradually increase the training set size. For sentence classification and natural language inference tasks, we randomly select a fixed test set of size 2000. For question answering, the full dev set is used for evaluation. 

\begin{table*}[t!]
\centering
\resizebox{\textwidth}{!}{
\begin{tabular}{l}
\hline
\textbf{Task} \\
\hline
Classfication \& NLI \\
QA \\
\hline
\end{tabular}
\begin{tabular}{ccccccccc}
\hline
\textbf{Batch Size} & $\boldsymbol{\alpha}$ & $\boldsymbol{\beta}$ & $\boldsymbol{\gamma}$ & $\boldsymbol{\eta}$ & $\boldsymbol{A_{1}}$ & $\boldsymbol{A_{2}}$ & \textbf{Adv. Step} & \textbf{Adv. LR} \\
\hline
8 & 0.3 & 0.3 & 10 &20& 0 & 10 & 3 & 2$e^{-2}$ \\
12 & 5 (2) & 5 (2) & 10 & 20 & 0.7  & 10 (2) &1 & 5$e^{-2}$ \\
\hline
\end{tabular}
}
\caption{Hyperparameters for \system, number in brackets are for large version pre-trained models}
\label{table:hyperparameters}
\end{table*}

\subsection{Implementation}
For a fair comparsion, We employ BERT\textsubscript{\begin{tiny}BASE\end{tiny}}, BERT\textsubscript{\begin{tiny}LARGE\end{tiny}}, RoBERTa\textsubscript{\begin{tiny}BASE\end{tiny}}, RoBERTa\textsubscript{\begin{tiny}LARGE\end{tiny}}, and BERT\textsubscript{\begin{tiny}BASE\end{tiny}} with TMix \cite{chen2020mixtext} \footnote{Since the original TMix is trained without attention mask, in our experiments, we add attention mask aligning with \system.} as strong baselines. \system is applied in two forms: \system without CAL optimization (denoted as \system*) and standard \system. All models are concatenated with a two-layer perceptron with $Tanh$ as activation on the top. We adopt the fourth mix strategy in \mixup for question answering in \ref{adjust}.
We report accuracy for sentence classification and natural language inference and EM/F1 for question answering.

We summarize test hyperparameters in Table \ref{table:hyperparameters}, and introduce the model-specific mixup candidate layers as follows:
$\{8, 9, 10\}$ (see Section \ref{sec:analysis} for detailed illustration) is used for BERT\textsubscript{\begin{tiny}BASE\end{tiny}}, $\{20, 21, 22\}$ for BERT\textsubscript{\begin{tiny}LARGE\end{tiny}}, $\{6, 7, 8\}$ for RoBERTa\textsubscript{\begin{tiny}BASE\end{tiny}}  and $\{17, 18, 19\}$ for RoBERTa\textsubscript{\begin{tiny}LARGE\end{tiny}}.   
During experiments, every trial is repeated 3-5 times. While for large versions of the pre-trained models, we also observe some unstable results as mentioned by \citet{devlin2019bert}, so the outliers are removed and the average accuracy is reported to reduce randomness. We also present other implementation details and hyperparameters in Appendices \ref{appendix: settings}, \ref{appendix: parms} and \ref{appendix: searching space}.

\subsection{Results}
\textbf{Performance of \system on classification and NLI}. As shown in Table \ref{tab:main_results_tab1}, it can be seen that \system achieves the best performance at the most settings across different pre-trained models. Improvements become more significant when the data size decreases, which is aligned with the analysis by \citet{yue2020interventional} that when the size of training data gets smaller, the impact of spurious bias increases. Besides, a more remarkable performance gain is observed on SNLI than that on IMDB which is relatively a simpler task. For instance, RoBERTa\textsubscript{\begin{tiny}BASE\end{tiny}} achieves 81.57 with only 10 samples per class. A simple task may be less influenced by spurious bias thus limit the improvement of \system. To verify our intuition, we remove the attention mask to increase the difficulty on IMDB, and the improvement on BERT\textsubscript{\begin{tiny}BASE\end{tiny}} for \system increases to 4.5, 2.9, 1.7, 1.0 percent for 10, 50, 250, 1000 number of class samples setting. As a result, we conclude that benefits given by \system are more obvious for a challenging task where spurious bias is serious. 

\vspace{2pt} \noindent 
\textbf{Performance of \system on question answering}. \system achieves the best performance nearly across all settings (Table \ref{tab:qa_result_tab1}). Similar to the trend in sentence classification and NLI, the improvement increases as the data size decreases. 
 Particularly, \system improves BERT\textsubscript{\begin{tiny}BASE\end{tiny}} by 10.88\%, 3.43\%, 2.08\% in EM respectively when data is 1/20, 1/10, 1/5 of the full SQuAD 1.1. Another noteworthy point is on average there are large improvements in EM than F1 (For example, 5.22\%, 4.11\% and 1.42\% in EM on SQuAD 2.0 and 4.35\%, 3.60\% and 1.30\% in F1 for BERT\textsubscript{\begin{tiny}BASE\end{tiny}} as data size grows), which shows that our mixup strategy can help model find answer boundary more precisely through the counterfactuals on non-answer context.
Evaluation results in Table \ref{tab:qa_all_result} shows \system still effective with full data.

\vspace{2pt} \noindent 
\textbf{Improvement of \system becomes more significant under adversarial QA}. We also observe a larger boosting on SQuAD 2.0 than SQuAD 1.1 averagely (Table \ref{tab:qa_all_result}), which demonstrates that \system can achieve more considerable improvement on more adversarial data than on benign data. Full data result also align with the trend (Table \ref{tab:qa_result_tab1}), which demonstrates the effectiveness of \system on full adversarial data.

\subsection{Case Study}
We further explore a concrete spurious bias, which is a statistical association between labels and some certain phrase in inputs, that has no real causation with labels. 
Take the phrase \textit{on a bench} on SNLI for instance, it logically has no causal effect on sentence pairs relations but may have statistical correlations with labels. To achieve that, we manually build a training set with 100 samples of which $10\%$ are with label entailment, $80\%$ with label contradiction and $10\%$ with label neutral. Such skewed distribution indicates a spurious bias between \textit{on a bench} and \textit{contradiction}. Then we build an equal-size test set, also equipped \textit{on a bench}, with the $40\%$, $20\%$ and $40\%$ proportion respectively. Hence test set is a out of distribution (OOD) set w.r.t. phrase \textit{on a bench}.
The result shows that BERT\textsubscript{\begin{tiny}BASE\end{tiny}}-\system improves the baseline significantly from 0.35 to 0.48 in Acc by successfully overcome the bias. For instance,
samples with phrase \textit{"a man sits on a bench"} are all contradictions during training but $83\%$ of them are neutral or entailment in test data. This phenomenon indicates that \system successfully alleviates this particular bias to better estimate the true causal effect between inputs and outputs, especially under OOD setting.

\begin{figure}[tb]
\setlength{\abovecaptionskip}{2pt}
 \centering
 \hspace{-0.8cm}
 \includegraphics[width=0.7\linewidth]{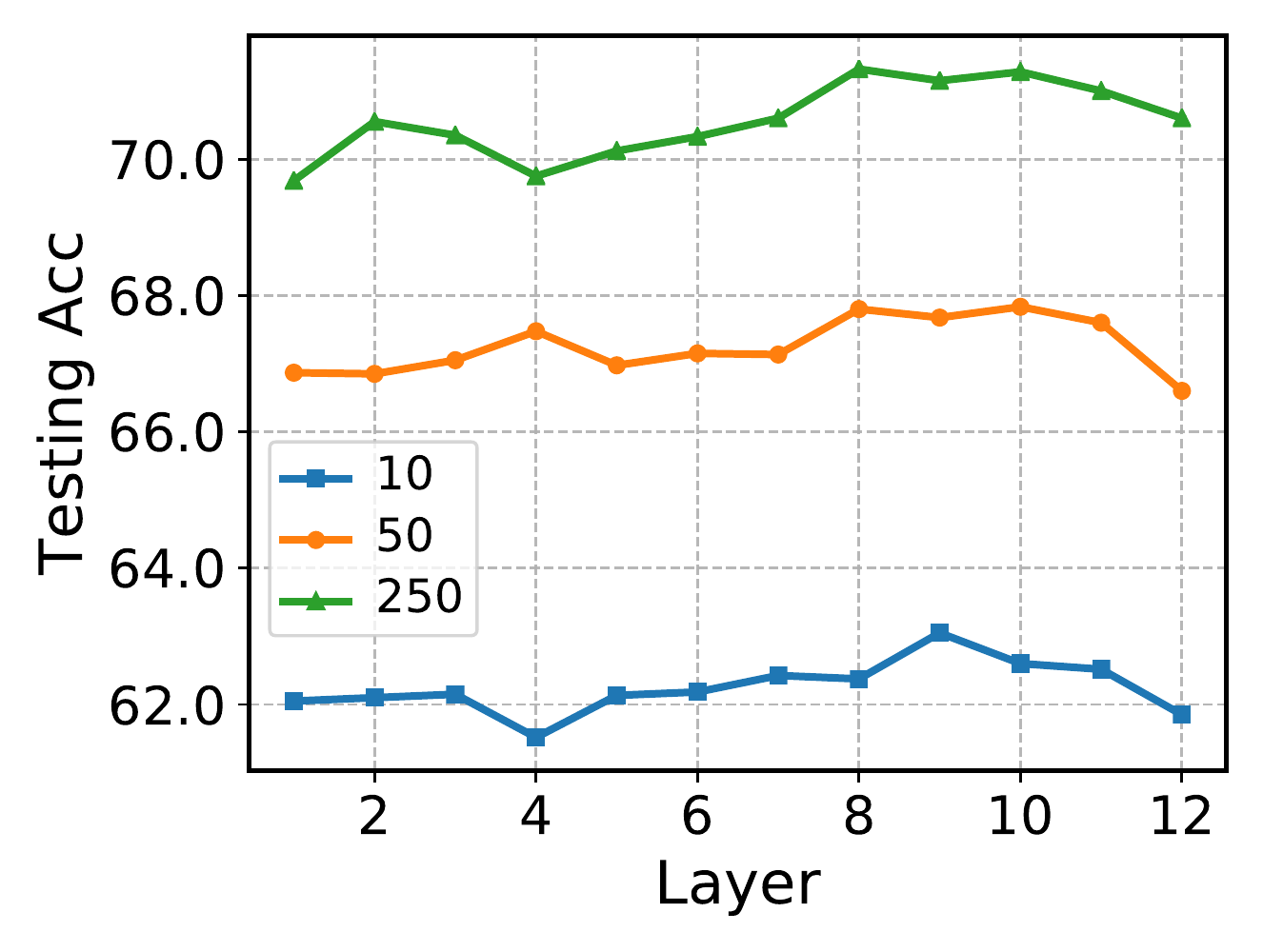}
 \caption{The impact of different interpolation layer on testing accuracy on Yahoo! Answer dataset.}
 \label{fig:layer_plot}
 \vspace{-0.5cm}
\end{figure}

\subsection{Analysis}\label{sec:analysis}
\textbf{Impact of Interpolation Layers}. We study the impact of using different layers for interpolation on BERT\textsubscript{\begin{tiny}BASE\end{tiny}} and the result is shown in Figure \ref{fig:layer_plot}. It is clear that the 8, 9, 10-th layers obviously outperform other layers consistently. In addition, we take $\{8, 9, 10\}$ layers as a candidate set and then sample one of them in each batch, which can further improve the model performance. 
For the other 3 language models, due to the limited computational resources, we experiment on several layers combinations and report the best one.

\vspace{2pt} \noindent 
\textbf{Performance and Stability of \system*}. It is observed that \system* can also improve pre-trained models but is less impressive than \system. More importantly, \system* is more unstable and sometimes even worse than baselines, e.g., \system* for RoBERTa\textsubscript{\begin{tiny}LARGE\end{tiny}} on IMDB and RoBERTa\textsubscript{\begin{tiny}LARGE\end{tiny}} on SQuAD 2.0. The reason can be seen in section \ref{sec:CRM}, where we discuss how CAL helps \system in both performance and stability. 


\begin{figure}[t]
\setlength{\abovecaptionskip}{2pt}
 \centering
 \includegraphics[width=1.0\linewidth]{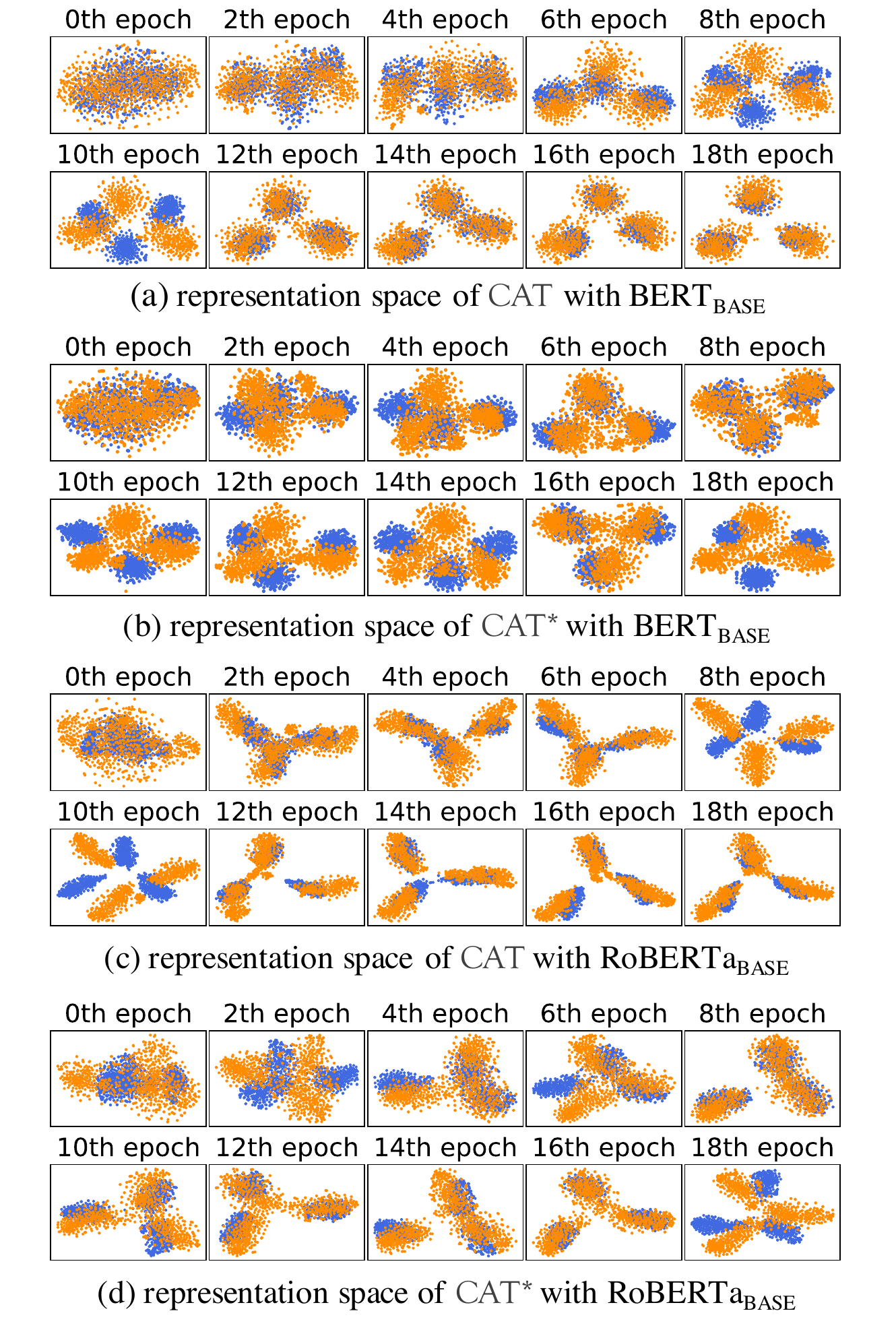}
\caption{Representation space visualization through tSNE for \system and \system*. during the training process on SNLI data with 250 samples per class. (a) and (b) represent \system and \system* on BERT\textsubscript{\begin{tiny}BASE\end{tiny}} and (c) and (d) for RoBERTa\textsubscript{\begin{tiny}BASE\end{tiny}}}.
\label{fig:scatter_plot_1}
\vspace{-0.55cm}
\end{figure}

\vspace{2pt} \noindent
\textbf{Training Process of \system and \system*}. We further explore how normal training data and counterfacutals evolve in latent space during training, particularly by visualization. Taking \system on SNLI as an example, the representations during training are shown in Figure \ref{fig:scatter_plot_1} (a), in which blue nodes denote original data representations and orange ones are counterfactuals. Obviously, We can conclude the process as three main stages illustrated as follows:





\vspace{2pt} \noindent
\indent \textit{First stage (epoch 0-4)}: The original data representations and counterfactuals are entangled together and the decision boundary for original data is not clear. 
 
\vspace{2pt} \noindent
\indent \textit{Second stage (epoch 6-10)}: Model begins to converge and the decision boundary for original data becomes more clear. Counterfactuals diverge gradually from the original data and clusters at the decision boundary. Such observation indicates that the classification loss manifold is not smooth enough and has many cliffs. Therefore, the gradient of CAL tends to pull the counterfactuals far away from the original data. In other words, $\gamma L(M^{(\theta)}(\tilde{h}^{(i)}), y^{(i)}))$ dominates the trade-off game over $\norm{\lambda^{(i)}}_p$. 
 
\vspace{2pt} \noindent
\indent \textit{Third stage (epoch 12-18)}: Counterfactuals move closer to the original data and few nodes are at the decision boundary. This observation shows the classification loss manifold becomes smoother and most of the cliffs disappear, thus main part of generated gradient of CAL during optimization starts to maximize $\norm{\lambda^{(i)}}_p$ instead of $\gamma L(M^{(\theta)}(\tilde{h}^{(i)}), y^{(i)}))$, which pulls counterfactuals close to the original data.
 




Through counterfactuals, \system alleviates the impact of spurious bias and encourages the model to discover the causal effect between representations and labels, thus helping pre-trained models construct a smoother and clearer loss manifold. The phenomenon that counterfactuals finally locates nearby the original data in representation place at stage 3 is well consistent with our counterfactual definition: minimally-different from original data but leads to different labels.

While for \system*, as shown in Figure \ref{fig:scatter_plot_1} (b), we also observe the similar stage 1 and stage 2, but the stage 3 is significantly different.
Such difference indicates \system* could only conduct random interpolation from a Beta distribution without CAL to further optimize counterfactuals. As a result, lots of interpolation representations will locate at the vicinity of the decision boundary when the model converges.
Such representation has two obvious weaknesses compared with that of \system: i) It may lead to the unstable performance of CRM since $\\Phi(M^{(\theta)}(\tilde{h}^{(i)}))$ could be extremely small. ii) The interpolated representation is not as efficient as minimal-different counterfactual ones in boosting performance as shown in \citet{kaushik2019learning}.


Additonally, we observe similar training process visualization when applying \system and \system* on SNLI for RoBERTa\textsubscript{\begin{tiny}BASE\end{tiny}}. The results are shown in Figure \ref{fig:scatter_plot_1} (c) and \ref{fig:scatter_plot_1} (d) respectively, which demonstrates an consistent convergence process has been achieved cross different pre-trained models.

\section{Conclusion}
To alleviate the spurious correlation bias in training corpus and encourage causal discovery instead of simple correlations, we propose \system from the causality perspective for introducing counterfactual representations in the training stage through latent space interpolation.
Through extensive experiments on three benchmarks on the text classification, natural language inference and question answering tasks, we demonstrate that \system is effective in promoting testing accuracy especially in the small data scenario, which outperforms SOTA baselines across different pre-trained models.

\section*{Acknowledgements}

We appreciate Hui Xue, Shi Chen and Jizhou Kang for sharing their pearls of wisdom. We also would like to express our special thanks of gratitude to the EMNLP anonymous reviewers for many helpful comments. This work was supported by Shining Lab and Alibaba Group through Alibaba Research Intern Program.

\bibliography{custom}
\bibliographystyle{acl_natbib}
\newpage
\appendix

\section{Appendices} \label{sec:appendix}

\subsection{Same Label Data Mixup} \label{appendix: same_label}
During \mixup, it could happen that $\tilde{y}^{(i)} = y^{(i)}$ especially if the two original samples used for \mixup are with the same label. In this case $\omega(h^{(i)}) \approx 1$ and CRM degenerate into ERM thus do no harm to our discriminative model.
Furthermore, since $\lambda^{(i)}$ is an exogenous random variable, as the training steps grow, we can expect a different $\tilde{h}^{(i)}$ in the next epoch such that $\tilde{y}^{(i)} \neq y^{(i)}$.

\subsection{Beta Distribution Hyperparameters}  \label{appendix: beta_pdf}
An interesting question here is the choice of prior $Beta(\alpha, \beta)$. 
Intuitively, the expectation of $Beta(\alpha, \beta) $ is $\alpha/(\alpha + \beta)$, thus one may want this value be close to 1 to make counterfactuals start searching from vicinity of observations. In experiments we find this strategy works better when the dataset is small. 
We think the reason is that when observations become more, the average distance between each sample goes down and the observation distribution become denser, thus the gains of searching from the vicinity is limited. 
Another noteworthy point is whether to choose a unimodal or bimodal distribution. Unimodal distribution generates $\lambda^{(i)}$ in the middle while bimodal distribution pushes $\lambda^{(i)}$ close to either $0$ or $1$. 
During the experiments we find the latter is usually better since it can explore a large range of interpolations.

\subsection{Other Implementation Details}\label{appendix: settings}
The datasets statistic are in Table \ref{table:datasets_stat}

\begin{table}[htp]
\centering 
\resizebox{\linewidth}{!}{
\begin{tabular}{c}
\hline
\textbf{Dataset} \\
\hline
Yahoo! Answers \\
IMDB \\
SNLI \\
SQuAD 1.1 \\
SQuAD 2.0 \\
\hline
\end{tabular}

\begin{tabular}{ccc}
\hline
\textbf{Classes} & \textbf{Test} & \textbf{Task} \\
\hline
10 & 2000  & classification \\
2 & 2000 & classification  \\
3 & 2000 & NLI \\
NA & 10570 & QA \\
NA & 11873 & QA \\
\hline
\end{tabular}
}
\caption{The details of test set used in our experiments.} 
\label{table:datasets_stat} 
\end{table}

We train sentence classification and natural language inference tasks on a single Tesla V100, question answering tasks on Tesla P100 and NVIDIA A100. We observe some performance discrepancy on question answering especially when data is small so we train all base models on Tesla P100 and all large models on NVIDIA A100 for consistency.

\subsection{Other Hyperparameters}\label{appendix: parms}

For sentence classification and NLI tasks, epoch is set as 30 when the number of samples in each category is less than 50, otherwise is set as 20. For question answering, epoch is set as 2 regardless of training data size. 

Regarding warm-up steps, for sentence classification and natural language inference tasks, we take the first epoch for warm-up training. For question and answering tasks, we choose from $\{300, 600, 1000\}$ steps to find the best. For learning rate, we choose from $\{1e-5, 2e-5\}$ for small data (number of samples in each class is less than 50) and fix the learning rate to $1e-5$ otherwise. Max sequence length is set as 128 for Yahoo! Answers and SNLI, 256 for IMDB and 384 for SQuAD 1.1 and SQuAD 2.0. 

During experiments we find that  BERT\textsubscript{\begin{tiny}LARGE\end{tiny}} and RoBERTa\textsubscript{\begin{tiny}LARGE\end{tiny}} are harder to train than corresponding base version and sometimes crash for long time training. Therefore we make some adjustments to mitigate it such as a smaller learning rate, tighter CRM bound $A_1$ and $A_2$, longer warm-up steps, etc.

\subsection{Hyperparameters Searching}\label{appendix: searching space}
We list the hyperparameters searching space here. For $\alpha$ and $\beta$ in beta distribution,  we iterate through $[0.1, 0.3, 0.5, 0.7, 0.9, 2, 5, 10]$ to find the best setting. For $\gamma$ and $\eta$, we keep $\eta = 2\gamma$ to reduce complexity and search $\gamma$ from $0$ to $20$. For CRM lower and upper bound, we gradually narrow our interval from $[0, 10]$ to $[0.9, 2]$ to find the optimal setting. We find that \system is difficult to converge with a lower bound near to $0$ under QA task. Finally, as regard to adversarial settings, we try learning step in $[1,3,5,10]$ and learning rate in $[2e-3, 5e-3, 2e-2, 5e-2]$. For the CRM learning rate, we generally set $2e-5$. except that we find for BERT\textsubscript{\begin{tiny}LARGE\end{tiny}} and  RoBERTa\textsubscript{\begin{tiny}LARGE\end{tiny}} on SNLI, when sample size per class is less than 50, a smaller learning rate $1e-5$ will lead to better and more stable performance.

\end{document}